\documentclass[slate]{bytedance}
\usepackage[toc,page,header]{appendix}

\usepackage{needspace}
\usepackage{placeins}
\usepackage{amsfonts}
\usepackage{amssymb}
\usepackage{tabularx}
\usepackage{listings}
\usepackage{xcolor}
\usepackage{cancel}

\usepackage{tabulary,multirow,xspace}
\usepackage{fixmath,mathtools,nicefrac,mmstyle}
\usepackage{subcaption}
\usepackage{caption}
\usepackage{wrapfig}
\usepackage[misc]{ifsym}
\usepackage{colortbl}
\usepackage{multicol}
\usepackage[most]{tcolorbox}
\usepackage{pifont}

\definecolor{codegreen}{rgb}{0,0.6,0}
\definecolor{codegray}{rgb}{0.5,0.5,0.5}
\definecolor{codepurple}{rgb}{0.58,0,0.82}
\definecolor{backcolour}{rgb}{0.95,0.95,0.92}
\definecolor{boxblue}{RGB}{57,89,163}
\definecolor{boxbluebg}{RGB}{230,237,250}

\lstdefinestyle{mystyle}{
    backgroundcolor=\color{backcolour},
    commentstyle=\color{codegreen},
    keywordstyle=\color{magenta},
    numberstyle=\tiny\color{codegray},
    stringstyle=\color{codepurple},
    basicstyle=\ttfamily\footnotesize,
    breakatwhitespace=false,
    breaklines=true,
    captionpos=b,
    keepspaces=true,
    numbers=none,
    numbersep=5pt,
    showspaces=false,
    showstringspaces=false,
    showtabs=false,
    tabsize=2
}
\definecolor{mygray1}{gray}{.95}
\definecolor{mygray2}{gray}{.9}
\definecolor{mygray3}{gray}{.95}

\newlength\savewidth
\newcolumntype{x}[1]{>{\centering\arraybackslash}p{#1pt}}

\DeclareTextFontCommand{\tablebf}{\bfseries}
\newenvironment{papertabular}[1]{%
  \let\textbf\tablebf 
  \fontsize{10.95}{12.6}\selectfont
  \setlength{\tabcolsep}{5pt}%
  \begin{tabular*}{\linewidth}{@{\extracolsep{\fill}}#1}%
}{\end{tabular*}}

\newcommand{\app}{\raise.17ex\hbox{$\scriptstyle\sim$}}

\usepackage{graphicx}
\usepackage{amssymb}
\usepackage{pifont}
\usepackage{floatrow}
\usepackage{amsmath} 
\usepackage{float}
\usepackage{wrapfig}
\usepackage{multirow}
\usepackage{tcolorbox}
\tcbuselibrary{breakable, skins, raster}
\usepackage{listings}
\usepackage{algorithm}
\usepackage{algorithmic}

\newcommand{\name}{LoopVAE}
\title{\name{}: Recurrent Depth Across Scales\\for Visual Tokenization}
\hypersetup{pdftitle={LoopVAE: Recurrent Depth Across Scales for Visual Tokenization},pdfauthor={Zhiying Lu}}

\newcommand{\paperauthor}{Zhiying Lu}
\author{
\centerline{
\paperauthor
}
}

\affiliation[]{University of Science and Technology of China}
\contribution[]{\protect\email{arieseirack@mail.ustc.edu.cn}}

\abstract{
Hierarchical visual tokenizers typically allocate different processing blocks to different spatial scales. We ask how much of this computation can use the same parameters. \textbf{\textit{LoopVAE}} reuses a scale- and loop-conditioned core within and across scales, while keeping resolution-changing transitions independent. A four-block core executes 28 block applications per encoder or decoder. On ImageNet-256, the 29M-parameter convolutional model reaches 0.28 rFID and 32.54 dB PSNR under an approximately 30-epoch two-stage training budget, using approximately 65\% fewer parameters than the 84M reference VAEs. A non-adversarial Transformer ablation with the same execution graph finds competitive PSNR and SSIM under global sharing, although unshared blocks improve LPIPS. Targeted loop interventions show that completing the trained recurrence improves reconstruction and that even small feature updates can have substantial downstream effects. Truncation also exposes output-range errors, distinguishing useful recurrent computation from reliable early exit. Runtime profiling reveals the execution tradeoff: fewer stored weights require more arithmetic and longer runtime in the tested configurations. With convolutional and Transformer operators and single- or multi-resolution latent interfaces, LoopVAE establishes recurrent depth across scales as a parameter-sharing design axis for visual tokenization.
}

\begin{document}
\maketitle

\section{Introduction}
\label{sec:intro}

\begin{figure}[t]
\centering
\includegraphics[width=\linewidth]{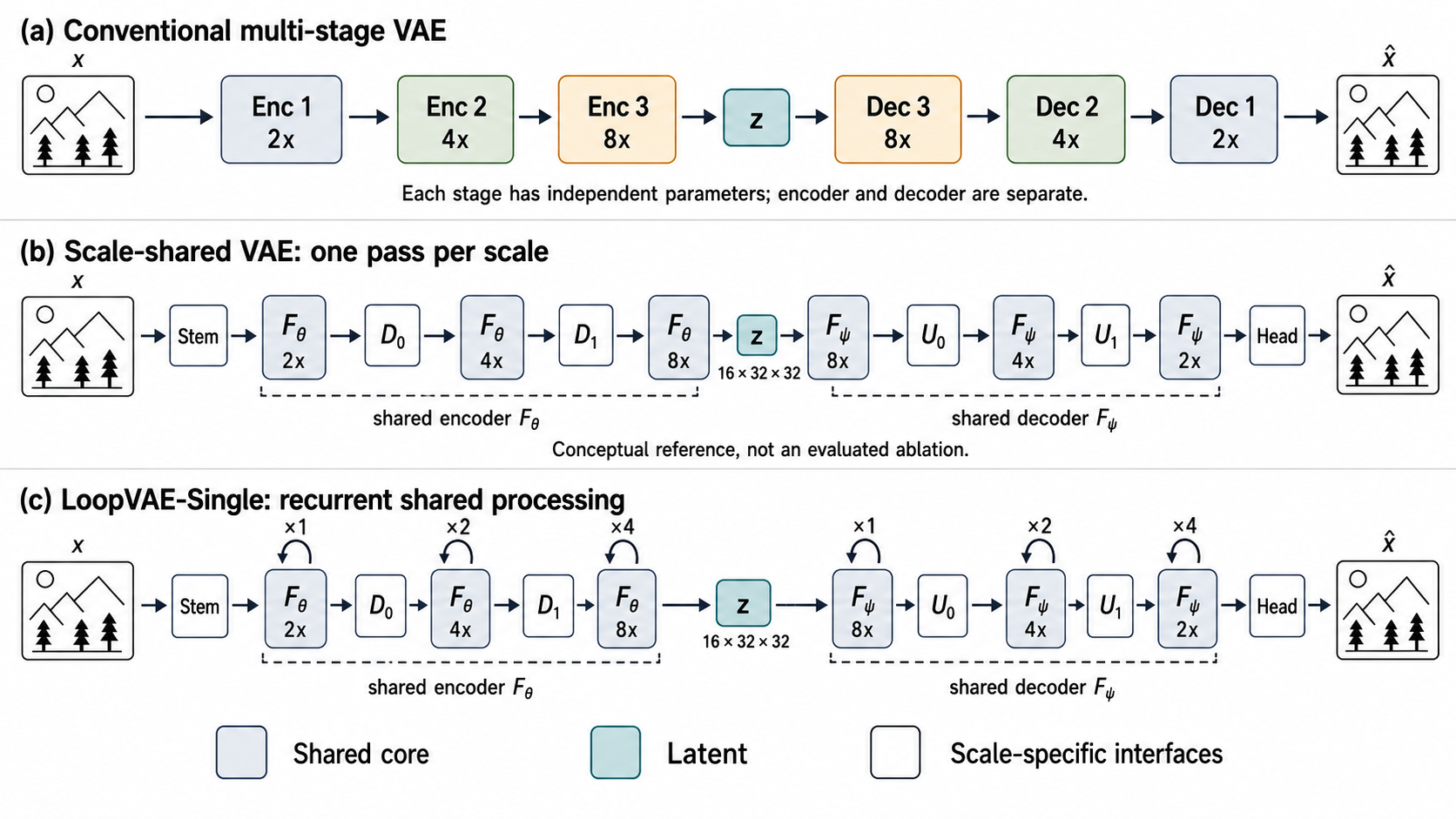}
\caption{\textbf{From stage-specific processing to LoopVAE-Single.} (a) Conventional stages own independent parameters. (b) A conceptual scale-shared reference uses one pass per scale; it is not an evaluated baseline. (c) LoopVAE-Single reuses a four-block stack with pass counts $[1,2,4]$ in each branch's traversal order. Encoder $F_\theta$ and decoder $F_\psi$ have separate weights, each shared across its own scales and loops. The stem, transitions, and output head remain scale specific; posterior heads and latent input projections are omitted for clarity. Image icons are schematic, not measured reconstructions.}
\label{fig:method_overview}
\end{figure}

Latent diffusion separates image generation into a visual tokenizer and a generative model over its latent representation~\cite{rombach2022ldm,peebles2023dit}. The tokenizer determines what information survives compression and how much spatial computation the generator receives. Recent work improves this interface through stronger compression~\cite{chen2025dcae}, Transformer scaling~\cite{hansen2025vitok}, semantic supervision~\cite{yao2025vavae,chen2025maetok,zheng2025rae}, and alternative reconstruction objectives~\cite{chen2025dito}. We study a complementary question: \emph{how should a hierarchical tokenizer allocate its processing parameters across depth and spatial scales?}

In conventional hierarchical autoencoders, changing resolution also changes the stack of blocks that processes the features. This couples tensor geometry to parameter identity. Recurrent-depth models separate executed depth from stored depth by repeatedly applying a shared core~\cite{dehghani2019universal,geiping2025recurrent}. Extending that principle across a visual hierarchy requires a shared update rule to operate on several feature resolutions, with separate transitions handling changes in geometry.

We introduce \name{}, an autoencoder organized around \emph{recurrent depth across scales}. Its encoder and decoder each own a small core reused at every spatial scale and loop step. Learned scale and loop embeddings condition residual updates, while a gated input-injection path supplies the feature that entered the current scale. Downsampling, upsampling, and interface projections retain independent parameters. Thus sharing applies to the fixed-resolution processing core, not to every component of the tokenizer. A four-block core with schedule $[1,2,4]$ executes 28 block applications per branch while storing only four core blocks.

This design raises three empirical questions. \textbf{Can global sharing retain reconstruction quality?} A 400k non-adversarial Transformer comparison holds the execution graph fixed while varying the number of stored core blocks from four to 12 or 28. Global sharing gives the highest PSNR and SSIM; fully unshared processing gives the lowest LPIPS. Separately, the final 29M-parameter CNN reaches 0.28 rFID and 32.54 dB PSNR on ImageNet-256 under an approximately 30-epoch two-stage training budget. These results motivate parameter sharing without implying that it improves every quality criterion.

\textbf{Does the trained recurrent depth remain necessary?} We examine a trained CNN through stage truncation and single-pass bypass interventions. Completing the trained schedule consistently improves reconstruction in the tested cases, and small relative feature changes can still have a substantial effect on the final output. At the same time, raw-output errors under truncation show that the recurrent trajectory is not automatically an anytime decoder. These complementary observations motivate studying both the utility of repeated computation and the calibration of intermediate states.

\textbf{What is the execution cost of parameter reuse?} Runtime profiling shows that the 8x looped CNN stores 65.3\% fewer parameters than a local Flux-style reference, but uses $1.57\times$ its estimated MACs and $2.85\times$ its batch latency. Reusing weights does not remove repeated high-resolution computation. This distinction between parameter economy and execution efficiency is central to interpreting the method.

Our contributions are: (i) a scale- and loop-conditioned autoencoder that separates shared processing from resolution transitions; (ii) evidence on sharing scope, depth sensitivity, and execution cost, including a trace-based analysis of small but consequential updates; and (iii) CNN and Transformer instantiations with standard single-resolution or selectable multi-resolution latent interfaces. Multi-resolution reconstruction and downstream diffusion demonstrate uses of these interfaces. Together, the results characterize where recurrent parameter sharing works, what its trained passes contribute, and which costs it leaves intact.

\section{Related Work}
\label{sec:related}

\paragraph{Continuous visual tokenizers.}
Variational autoencoders~\cite{kingma2014vae} provide a probabilistic continuous latent representation.  Perceptual and adversarial reconstruction objectives, also used in discrete tokenizers such as VQGAN~\cite{esser2021taming}, help preserve visual detail; latent diffusion models use perceptually trained autoencoders as their first stage~\cite{rombach2022ldm}.  Recent work improves high-compression tokenization with residual autoencoding~\cite{chen2025dcae}, studies Transformer tokenizer scaling~\cite{hansen2025vitok}, replaces a compound reconstruction recipe with a diffusion objective~\cite{chen2025dito}, or shapes latents with semantic supervision~\cite{yao2025vavae,chen2025maetok,zheng2025rae}.  These methods primarily change the objective, bottleneck, or source of representation supervision.  LoopVAE studies an orthogonal architectural variable: whether the same fixed-resolution operator can be reused throughout the spatial hierarchy.

\paragraph{CNN and Transformer autoencoders.}
Convolutional autoencoders encode locality and translation equivariance, while Vision Transformers~\cite{dosovitskiy2021vit} provide content-dependent global interaction.  ConvNeXt~\cite{liu2022convnext} modernizes the residual CNN design with large depthwise kernels and expanded channel mixing; ViTok~\cite{hansen2025vitok} studies how Transformer autoencoders scale for reconstruction and generation.  Window attention~\cite{liu2021swin} reduces the quadratic cost of high-resolution interaction.  We treat the spatial mixer as an implementation choice inside the same recurrent hierarchy.  The central question is sharing scope, not whether CNNs or Transformers are universally preferable.

\paragraph{Weight tying and recurrent depth.}
Universal Transformers~\cite{dehghani2019universal} reuse weights across depth, and looped Transformers can express iterative computation with a shallow shared network~\cite{giannou2023looped}.  Geiping et al.~\cite{geiping2025recurrent} train recurrent-depth language models with variable recurrence and study additional computation at test time.  SMELT~\cite{wang2026smelt} studies middle-layer looping in MoE Transformers while closely matching per-token FLOPs, stored parameters, and KV cache.  These works distinguish executed depth from independently parameterized depth.  LoopVAE studies a visual autoencoder whose core is tied across both recurrent steps and changing spatial resolutions.  Our fixed-schedule experiments establish neither test-time depth extrapolation nor a compute-matched advantage.  Scale conditioning identifies the resolution; its independent benefit remains to be isolated.

\paragraph{Recurrence in visual compression and generation.}
Recurrent image compression predates LoopVAE: Toderici et al.~\cite{toderici2016recurrent} use recurrent encoders and decoders with quantization to support variable-rate compression. Their progressive coding objective differs from our continuous latent interface and reuse of a core across a spatial hierarchy. Elastic Looped Transformers~\cite{goyal2026elt} reuse blocks in visual generative backbones and use intra-loop self-distillation to improve intermediate predictions. LoopVAE instead studies the tokenizer; our truncation diagnostics highlight why intermediate-output quality should not be inferred from weight sharing alone. The distinction is therefore the location and scope of recurrence, not the introduction of recurrent computation to vision.

\paragraph{Multi-resolution latent representations.}
Compression ratio controls both reconstruction capacity and the token sequence presented to a generative model.  Deep compression autoencoders~\cite{chen2025dcae} reduce spatial sequence length aggressively, whereas semantic or high-dimensional tokenizers preserve richer features at a larger latent cost~\cite{yao2025vavae,zheng2025rae}.  Most systems expose one bottleneck per trained tokenizer.  \name{}-Multi instead attaches several posterior heads to one shared hierarchy and supplies corresponding decoder entry points.  It therefore treats latent resolution as a selectable interface of one model, while leaving a controlled comparison against separately trained tokenizers to future work.

\Needspace{6\baselineskip}
\section{Method}
\label{sec:methodology}

\subsection{From Scale-Specific Stages to a Shared Operator}

Let $x \in \mathbb{R}^{3\times H\times W}$ be an image, $E$ an encoder, and $D$ a decoder.  A continuous tokenizer produces a Gaussian posterior and reconstructs a sample from it:
\begin{equation}
  q(z\mid x)=E(x), \qquad z\sim q(z\mid x), \qquad \hat{x}=D(z).
  \label{eq:vae}
\end{equation}
A conventional hierarchical encoder can be written as
\begin{equation}
  h_{s+1}=T_s\bigl(F_s(h_s)\bigr),
  \label{eq:stagewise}
\end{equation}
where $F_s$ processes features at scale $s$ and $T_s$ changes their resolution.  Both modules depend on the scale.  LoopVAE keeps $T_s$ scale specific but replaces the collection $\{F_s\}$ with one conditioned operator $F_\theta$.

The encoder and decoder use separate parameters, shown as $F_\theta$ and $F_\psi$ in Figure~\ref{fig:method_overview}, because analysis and synthesis solve different transformations.  Within either branch, however, the same operator is reused across every scale and loop step.  We write $F_\theta$ generically below for either branch's operator.  This separation defines the scope of our claim: we share the expressive fixed-resolution computation, while retaining independent downsampling, upsampling, input projection, and output projection modules wherever tensor geometry changes.

\subsection{Scale-Conditioned Depth Recurrence}

At scale $s$, let $h_{s,0}$ be the feature produced by the preceding transition.  A learned scale embedding $e_s$ identifies the resolution and a learned loop embedding $e_t$ identifies the current refinement step:
\begin{equation}
  c_{s,t}=e_s+e_t.
  \label{eq:condition}
\end{equation}
The feature is then updated $L_s$ times:
\begin{equation}
  h_{s,t+1}=F_\theta\!\left(
    h_{s,t}+\mathbf{1}_{t>0} I_s(h_{s,0}),\; c_{s,t}
  \right),
  \quad t=0,\ldots,L_s-1.
  \label{eq:loop}
\end{equation}
$I_s(h)=g_sP_s(h)$ is a scale-specific projection multiplied by a learned scalar $g_s$ initialized to zero.  $P_s$ is a $1\!\times\!1$ convolution for CNN features and a token-wise linear map for Transformer features.  The first iteration omits this injection.  On later iterations, it is added once before the entire $K$-block core, not before every block.  This construction provides an information shortcut; its independent benefit has not yet been isolated experimentally.

For the three-scale model, a stack of $K=4$ blocks is evaluated with $L=[1,2,4]$ in traversal order: 2x, 4x, and 8x for the encoder, but 8x, 4x, and 2x for the decoder.  Thus the decoder performs more loops at its finer scales, not at the same spatial scales as the encoder.  Each branch stores four core blocks but performs
\begin{equation}
  d_{\mathrm{core}}=K\sum_s L_s=4(1+2+4)=28
  \label{eq:block_calls}
\end{equation}
core block applications in each forward pass through the encoder or decoder.  Here $K$ is the stored core depth, $L_s$ is the recurrent pass count at scale $s$, and $d_{\mathrm{core}}$ is the executed core depth, excluding transitions and heads.  The same core therefore realizes depth through loops within each scale and weight reuse across scales.  This count is not a FLOP estimate: a block's cost depends on its spatial resolution.  All reported checkpoints use fixed loop schedules; additional test-time recurrence has not been evaluated.

\paragraph{Stored capacity and executed work.}
For $S$ scales and $p$ parameters per core block, the core parameter counts under global, scale-wise, and fully unshared processing are
\begin{equation}
 P_{\rm global}=Kp,\qquad P_{\rm scale}=SKp,\qquad
 P_{\rm unshared}=Kp\sum_s L_s.
 \label{eq:sharing_capacity}
\end{equation}
These counts exclude the transitions, heads, embeddings, and injection modules. They describe core capacity, not total-model compression ratios. For the same execution graph, all three designs apply $K\sum_s L_s$ blocks, with core arithmetic approximately $K\sum_s L_s\mathcal{C}(H_s,W_s,C)$, where $\mathcal{C}$ is one block's cost. Sharing changes which weights are read, not how often a block is executed. This separates the architectural variable tested by the sharing ablation from the runtime effects measured in Section~\ref{sec:resources}.

\subsection{Operator Instantiations}

\begin{figure}[p]
\centering
\includegraphics[height=0.88\textheight,width=\linewidth,keepaspectratio]{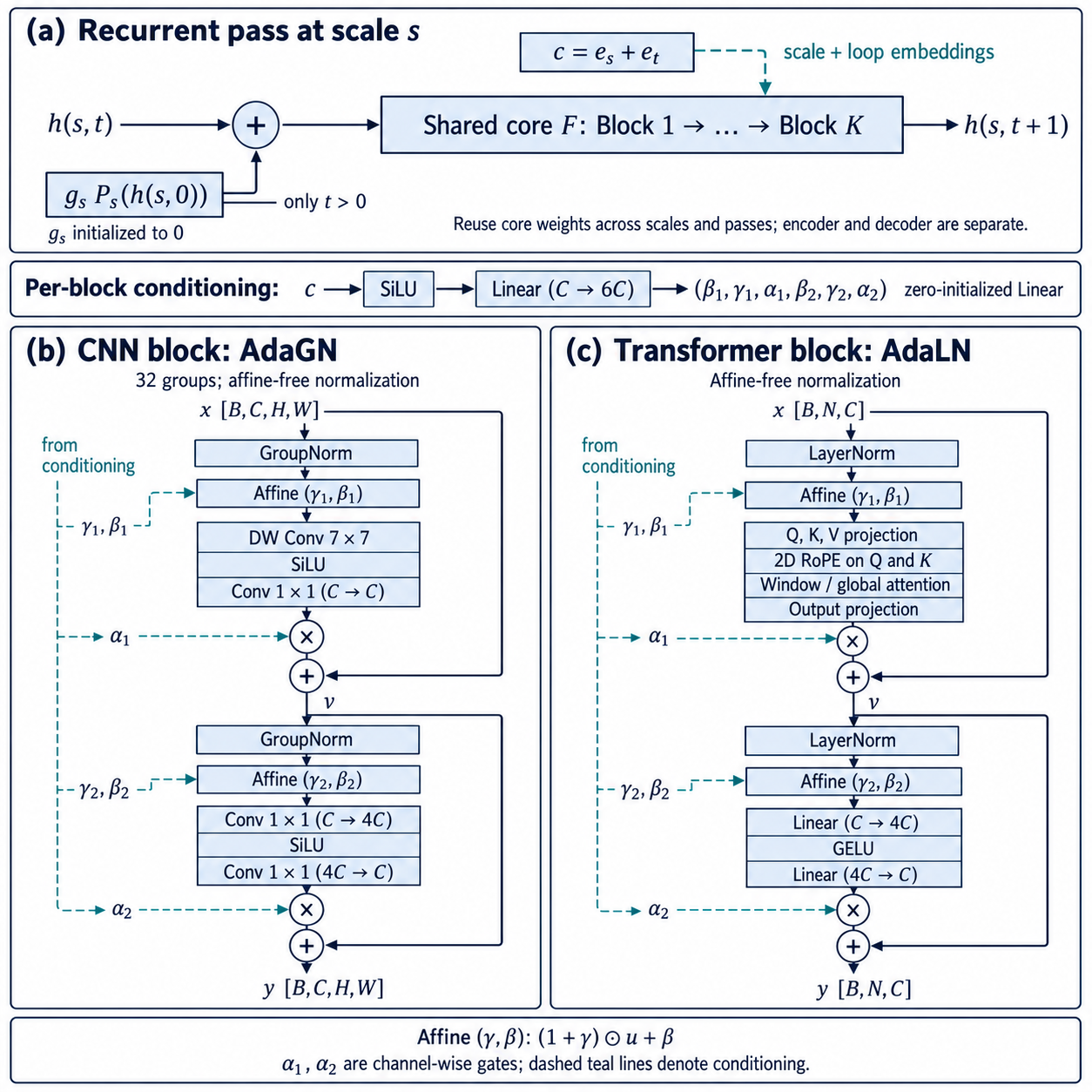}
\caption{\textbf{Inside the recurrent core.} Input injection precedes the entire core on passes $t>0$. Each distinct block has its own zero-initialized modulation projection. CNN blocks use AdaGN and SiLU; Transformer blocks use AdaLN, attention, and a GELU MLP. Both alternatives have two gated residual branches. The same core weights are reused across scales and passes, separately in each encoder and decoder.}
\label{fig:block_internals}
\end{figure}

Figure~\ref{fig:block_internals} details the recurrent wrapper and both block alternatives.  Scale and loop embeddings are learned lookup vectors; their sum conditions each block rather than being directly added to image features.  For block $j$, a SiLU followed by a linear projection produces six channel-wise vectors:
\begin{equation}
 (\beta_1,\gamma_1,\alpha_1,\beta_2,\gamma_2,\alpha_2)
 = W_j\,\mathrm{SiLU}(c_{s,t})+b_j.
 \label{eq:modulation}
\end{equation}
The projection maps width $C$ to $6C$ and starts with zero weights and bias.  Each block has its own projection, reused whenever that block executes.  The two residual updates are
\begin{align}
 v &= u+\alpha_1\odot M_1\bigl((1+\gamma_1)\odot N_1(u)+\beta_1\bigr),\nonumber\\
 y &= v+\alpha_2\odot M_2\bigl((1+\gamma_2)\odot N_2(v)+\beta_2\bigr).
 \label{eq:block_update}
\end{align}
Here $N_i$ is affine-free normalization and $M_i$ is a spatial or channel mixer.  Modulation vectors broadcast over spatial positions or tokens; $\alpha_i$ are unconstrained channel-wise gates, not sigmoid probabilities.  Zero-initialized gates make each block initially an identity.  They are distinct from the scalar input-injection gate $g_s$.

\paragraph{Convolutional operator.}
Our primary model uses a conditioned ConvNeXt-style block~\cite{liu2022convnext}.  Both $N_i$ are GroupNorm with 32 groups and $\epsilon=10^{-6}$, so this is \emph{AdaGN}, not AdaLN.  $M_1$ is a depthwise $7\!\times\!7$ convolution, SiLU, and a $1\!\times\!1$ projection.  $M_2$ is a $1\!\times\!1$ expansion from $C$ to $4C$, SiLU, and a $1\!\times\!1$ projection back to $C$.  The main configuration uses width 384 and zero dropout.

\paragraph{Transformer operator.}
The same hierarchy can instead use a DiT-style self-attention block~\cite{peebles2023dit}.  Here $N_i$ are affine-free LayerNorm with $\epsilon=10^{-6}$, yielding AdaLN.  $M_1$ projects queries, keys, and values, applies two-dimensional rotary embeddings~\cite{su2024roformer} to queries and keys, performs attention, and projects the output.  At high-resolution scales, non-overlapping local windows~\cite{liu2021swin} limit attention cost.  $M_2$ is a $C\rightarrow4C\rightarrow C$ linear MLP with GELU.  These blocks receive scale and recurrence indices, not diffusion timesteps or class labels.

\subsection{Single- and Multi-Resolution Tokenizers}

\paragraph{LoopVAE-Single.}
The single-resolution encoder uses a stride-2 stem followed by the three scales $2\mathrm{x}\rightarrow4\mathrm{x}\rightarrow8\mathrm{x}$.  A final head predicts the mean and log variance of a diagonal Gaussian.  At $256\!\times\!256$, the \texttt{f8d16} latent has shape $16\times32\times32$.  The decoder projects this latent to the shared width and traverses the scales in reverse.  Its core blocks are independent of the encoder blocks but are reused in the same way.  Consequently, LoopVAE-Single can replace a conventional \texttt{f8d16} VAE without changing the downstream diffusion input shape.

\begin{figure}[t]
\centering
\includegraphics[width=\linewidth]{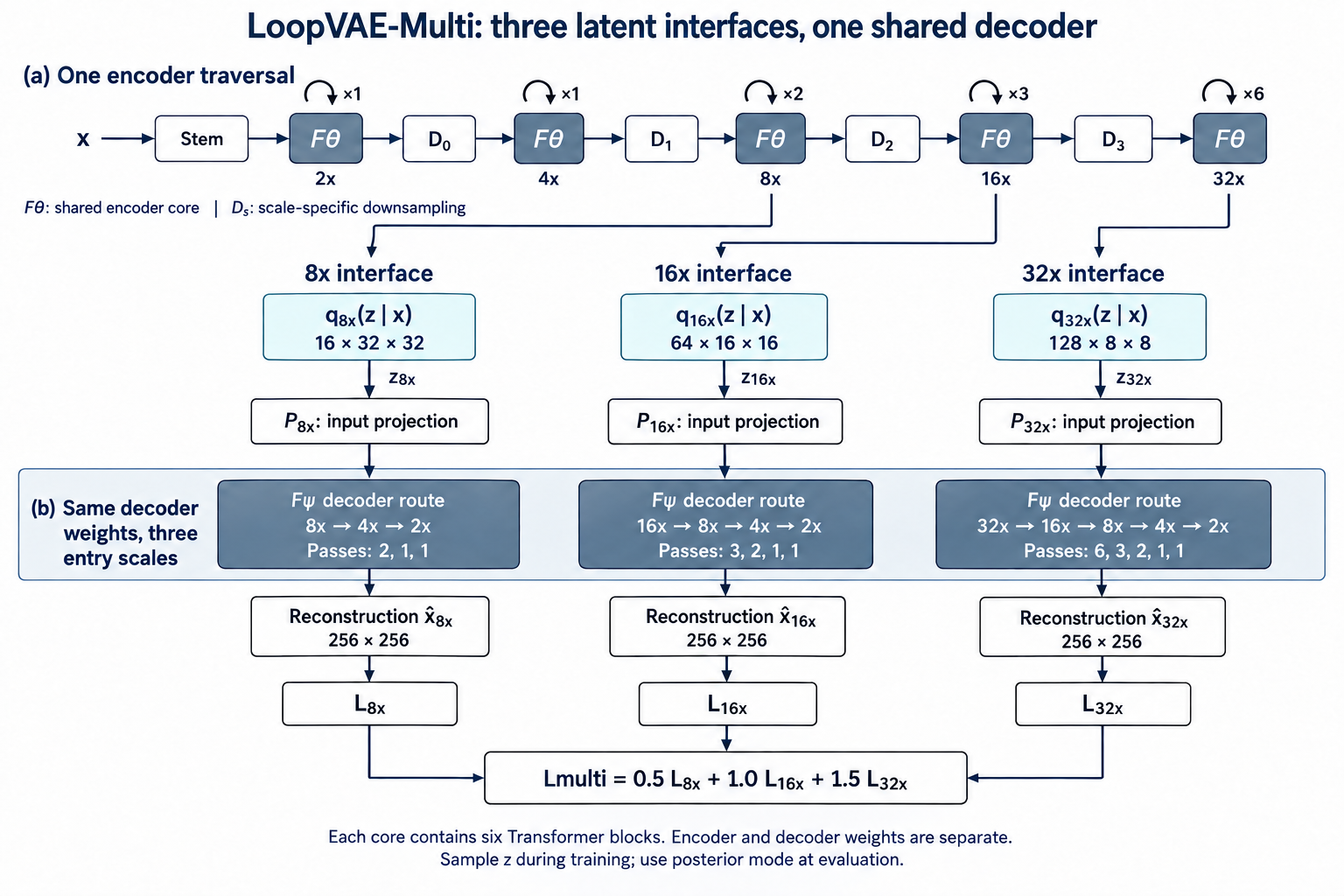}
\caption{\textbf{LoopVAE-Multi and joint training.} A single encoder traversal produces three independent posterior heads. Each sampled latent is decoded separately through its own input projection and the corresponding suffix of one shared decoder. The three decoder boxes depict executions with the same weights, not independently parameterized decoders or a concatenation of latents. Scale labels denote spatial downsampling. Each $F$ contains six Transformer blocks; the per-interface losses include reconstruction, KL, and active adversarial terms. Their weighted sum jointly trains the encoder and decoder.}
\label{fig:multi_overview}
\end{figure}

\paragraph{LoopVAE-Multi.}
Figure~\ref{fig:multi_overview} shows the multi-resolution model, which extends the hierarchy to $2\mathrm{x}\rightarrow4\mathrm{x}\rightarrow8\mathrm{x}\rightarrow16\mathrm{x}\rightarrow32\mathrm{x}$.  Independent posterior heads are attached at the last three scales:
\begin{equation}
  \{q(z_{8\mathrm{x}}\mid x),q(z_{16\mathrm{x}}\mid x),q(z_{32\mathrm{x}}\mid x)\}=E_{\mathrm{multi}}(x).
  \label{eq:multi}
\end{equation}
Their channel dimensions are 16, 64, and 128, giving latent shapes $16\times32^2$, $64\times16^2$, and $128\times8^2$ for a $256^2$ image.  Here 8x, 16x, and 32x denote spatial downsampling, not bitrate: the first two interfaces both contain 16,384 scalars, while the last contains 8,192.  Each head has its own normalization and output projection.  The decoder similarly has one input projection per latent shape, enters the shared synthesis hierarchy at the selected scale, and executes only the remaining scales.  The exploratory five-scale configuration uses six Transformer blocks and encoder loop schedule $[1,1,2,3,6]$.

\subsection{Training Objective}

Training has two stages. Stage 1 is non-adversarial pretraining with a reconstruction, perceptual, and KL objective:
\begin{equation}
  \mathcal{L}_{\mathrm{rec}}
  =\lVert x-\hat{x}\rVert_1
  +\lambda_{p}\mathcal{L}_{\mathrm{LPIPS}}(x,\hat{x})
  +\lambda_{\mathrm{KL}}D_{\mathrm{KL}}\!\left(q(z\mid x)\,\|\,\mathcal{N}(0,I)\right).
  \label{eq:loss_rec}
\end{equation}
In the implementation, reconstruction terms are averaged over pixels and the KL term is averaged over latent elements and batch examples.  During training, the single-latent model additionally perturbs posterior samples with zero-mean Gaussian noise of standard deviation 0.05; evaluation uses the posterior mode without this perturbation.
By default, Stage 1 runs for 600k steps. Stage 2 initializes the autoencoder from the Stage-1 weights and continues training with a hinge adversarial loss to sharpen high-frequency detail:
\begin{equation}
  \mathcal{L}=\mathcal{L}_{\mathrm{rec}}+\lambda_{a}\mathcal{L}_{\mathrm{adv}}.
  \label{eq:loss_adv}
\end{equation}
Here $\mathcal{L}_{\mathrm{adv}}$ denotes the generator's negative mean discriminator score; the discriminator is trained with the hinge objective.  The weight $\lambda_a$ uses a clipped, gradient-norm-based adaptive coefficient after the configured activation threshold.  The main CNN run uses frozen DINOv2 features~\cite{oquab2024dinov2} with a learned head, differentiable augmentation inspired by DiffAugment~\cite{zhao2020diffaugment}, and lazy R1 regularization~\cite{mescheder2018training}.  The multi-resolution run instead uses a convolutional discriminator; its per-interface reconstruction, KL, and active adversarial losses are combined with fixed scale weights.

\FloatBarrier
\section{Experiments}
\label{sec:experiments}

We evaluate reconstruction quality, sharing scope, recurrent-depth sensitivity, and execution cost. Multi-resolution reconstruction and downstream diffusion test additional uses of the learned interfaces. Checkpoint evaluations, within-checkpoint interventions, and architectural profiling address complementary questions; their configurations are specified separately.

\subsection{Setup}

\paragraph{Data and metrics.}
The reconstruction protocol uses the ImageNet-1k training split and the 50k-image validation split at $256\!\times\!256$~\cite{deng2009imagenet}.  We report rFID~\cite{heusel2017fid}, PSNR, LPIPS~\cite{zhang2018lpips}, and SSIM~\cite{wang2004ssim}, with posterior-mode decoding.  The sharing ablation is evaluated before adversarial training; its table deliberately reports only paired reconstruction metrics.  Results are individual checkpoint evaluations, not averages across seeds.

\paragraph{Training stages and budgets.}
Stage 1 is non-adversarial pretraining for 600k steps by default; Stage 2 continues from these weights with GAN training. Stage-2 step labels count only the second stage and exclude Stage 1. For the main reconstruction comparison, we report a nominal total budget of approximately 30 epochs: approximately 15 epochs of Stage 1 plus 15 epochs of Stage 2. The reference configuration uses eight GPUs, each with 96\,GB of device memory, and four images per GPU, giving a global batch size of 32. Appendix~\ref{sec:training_budget} specifies the epoch-estimation convention.

\paragraph{Models and checkpoint scope.}
The primary CNN LoopVAE-Single uses width 384, four core blocks per branch, loop schedule $[1,2,4]$, and an \texttt{f8d16} latent. It stores 29.07M encoder and decoder parameters, excluding perceptual and discriminator networks. Table~\ref{tab:single_results} reports the CNN and ViT results under this two-stage budget at \texttt{f8d16}: eightfold spatial downsampling and 16 latent channels. The sharing ablation evaluates the V3 Transformer at the non-adversarial Stage-1 400k checkpoint, before the default pretraining endpoint. Loop interventions and downstream diffusion use the Stage-2 610k CNN tokenizer, while all three Multi interfaces are evaluated jointly at Stage 2, step 220k. These auxiliary evaluations are kept distinct from the final reconstruction comparison.

\subsection{Reconstruction Results}

We compare the \texttt{f8d16} models with the SD3~\cite{esser2024sd3} and FLUX~\cite{blackforest2024flux} VAEs, which expose the same latent shape at $256^2$ resolution.

\begin{table}[htbp]
\centering
\caption{\texttt{f8d16} reconstruction on ImageNet-256. Our LoopVAE models use approximately 30 epochs in total: 15 without GAN (Stage 1) followed by 15 with GAN (Stage 2). Baseline training recipes are not matched. Bold denotes column-best results, including ties at the reported precision.}
\label{tab:single_results}
\begin{papertabular}{lccccc}
\toprule
Model (\texttt{f8d16}) & Params $\downarrow$ & rFID $\downarrow$ & PSNR $\uparrow$ & LPIPS $\downarrow$ & SSIM $\uparrow$ \\
\midrule
SD3 VAE & 84M & 0.19 & 31.29 & 0.060 & 0.88 \\
FLUX-VAE & 84M & \textbf{0.18} & \textbf{32.80} & \textbf{0.044} & \textbf{0.91} \\
\midrule
LoopVAE-ViT (ours) & 34M & 0.32 & 31.43 & 0.072 & 0.89 \\
LoopVAE-CNN (ours) & \textbf{29M} & 0.28 & 32.54 & 0.048 & \textbf{0.91} \\
\bottomrule
\end{papertabular}
\end{table}

Under this approximately 30-epoch two-stage budget, the 29M-parameter LoopVAE-CNN reaches 0.28 rFID, 32.54 PSNR, 0.048 LPIPS, and 0.91 SSIM (Table~\ref{tab:single_results}).  It uses approximately 65\% fewer parameters than the 84M baselines and matches FLUX-VAE's displayed SSIM, while FLUX-VAE has better rFID, PSNR, and LPIPS.  The 34M-parameter LoopVAE-ViT reaches 0.32 rFID.  Training recipes are not controlled, so quality differences cannot be attributed solely to recurrence or operator choice.

\subsection{Non-Adversarial Sharing Ablation}

The V3 Transformer ablation compares three sharing scopes at the Stage-1 400k checkpoints without GAN training. Each follows the same three-scale loop graph with 28 core block applications per encoder or decoder. Global sharing stores four blocks per branch; scale-wise sharing stores 12, reusing each scale's blocks across its loop steps; fully unshared processing stores 28. Thus the comparison varies stored core capacity while retaining the intended sequence of block applications.

\begin{table}[htbp]
\centering
\caption{V3 Transformer sharing ablation at Stage 1, step 400k, without adversarial training. Stored blocks are counted per branch; all designs execute 28 core block applications per branch. rFID is omitted from this non-GAN comparison. Bold marks the best value in each metric.}
\label{tab:sharing_designs}
\begin{papertabular}{lrrrr}
\toprule
Sharing rule & Stored blocks & PSNR $\uparrow$ & LPIPS $\downarrow$ & SSIM $\uparrow$ \\
\midrule
Global & 4 & \textbf{31.163} & 0.0852 & \textbf{0.8875} \\
Scale-wise & 12 & 30.412 & 0.0841 & 0.8713 \\
Fully unshared & 28 & 31.043 & \textbf{0.0797} & 0.8852 \\
\bottomrule
\end{papertabular}
\end{table}

Global sharing has the highest PSNR and SSIM, exceeding fully unshared processing by 0.120 dB and 0.0023, respectively, while using one seventh as many stored core blocks. Fully unshared processing achieves the lowest LPIPS, improving over global sharing by 0.0055. Scale-wise sharing lies between them in LPIPS but has the lowest PSNR and SSIM. Removing sharing therefore does not uniformly improve reconstruction: the smallest shared core remains competitive on paired fidelity, while unshared capacity benefits LPIPS. These single-run results do not quantify seed variation or isolate possible effects of data ordering.

\subsection{Recurrent Depth and Loop-Position Sensitivity}
\label{sec:loop_dynamics}

Does repeated application of the shared core contribute useful computation, or are some passes effectively redundant? We examine the trained CNN tokenizer through two inference-time interventions: truncating a single stage to its first $k$ passes, and bypassing one complete pass while retaining all subsequent loop indices. These tests expose sensitivity to recurrent depth and position without retraining the model.

\paragraph{Controlled interventions.}
We use 16 typical test images from TokBench~\cite{wu2025tokbench} for a targeted case study of fine visual detail, with particular attention to faces and text. The experiments use the Stage-2 610k \texttt{f8d16} CNN checkpoint, corresponding to approximately 15 nominal epochs of GAN training after Stage 1, with posterior-mode decoding in bfloat16. We measure how recurrent interventions affect paired reconstruction, rather than running the full TokBench text-recognition and face-similarity evaluation. All selected images enter the reported statistics; their identities and selection order are documented in Appendix~\ref{sec:loop_protocol}. Encoder stages at f2/f4/f8 execute 1/2/4 passes; decoder stages at f8/f4/f2 execute 1/2/4. Here f2 denotes an internal half-resolution feature map, not a latent compression ratio.

For a prefix intervention, only the selected stage is shortened; the remaining network completes its trained schedule. Decoder interventions use the same full-encoder latent. For a bypass, we replace the entire pass output by its pre-injection state, removing both injection and core update while preserving later loop indices. These are counterfactual final reconstructions, not native intermediate RGB predictions or independently trained shallow models. All 96 full-prefix reconstructions (six stages on 16 images) match their corresponding baseline arrays exactly. Appendix~\ref{sec:loop_protocol} gives the measurement protocol.

\begin{figure}[!t]
  \centering
  \includegraphics[width=\linewidth]{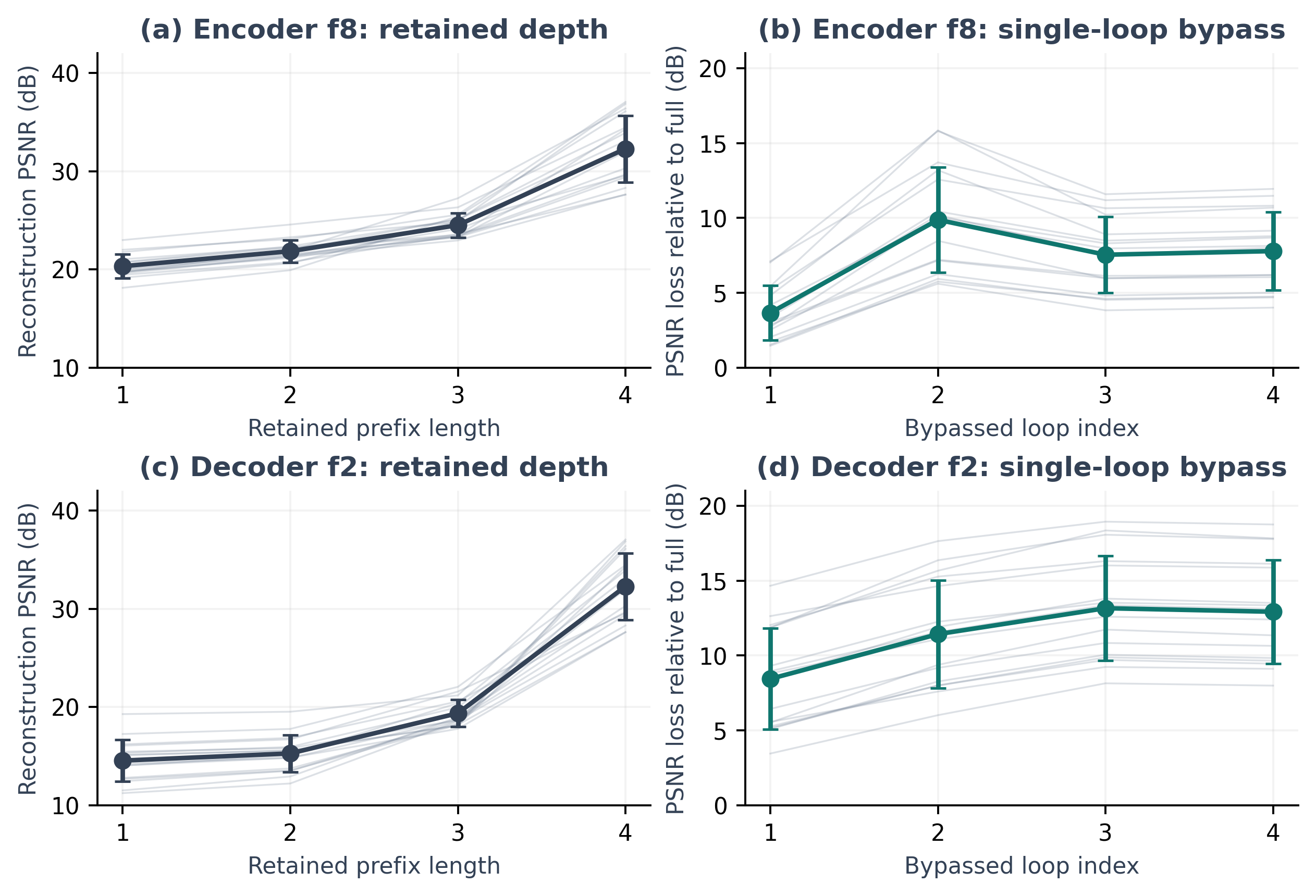}
  \caption{\textbf{Recurrent depth and position affect reconstruction.} Left: truncating one stage while keeping all other stages fixed. Right: independently bypassing one loop; positive values denote PSNR loss relative to the full schedule. Thin lines show the 16 test images; markers and bars show the mean and sample standard deviation across images, not seed uncertainty. The second encoder f8 pass and third decoder f2 pass are the most bypass-sensitive positions in their respective stages on every tested image.}
  \label{fig:loop_sensitivity}
\end{figure}

\paragraph{Depth benefits persist across images.}
At encoder f8, mean PSNR increases from 20.32 to 21.85, 24.50, and 32.28 dB as the retained prefix grows from one to four passes (Figure~\ref{fig:loop_sensitivity}). At decoder f2, the corresponding values are 14.54, 15.26, 19.36, and 32.28 dB. All four multi-pass stages---encoder f4/f8 and decoder f4/f2---have non-decreasing PSNR and SSIM on \emph{every} evaluated image. The pattern also holds for PSNR computed before output clipping. Thus the mean trend does not conceal a counterexample within this subset. Full-schedule reconstruction averages 32.279 dB PSNR and 0.9424 SSIM; these subset metrics are not substituted into the main reconstruction table.

\begin{figure}[p]
  \centering
  \includegraphics[width=.94\linewidth]{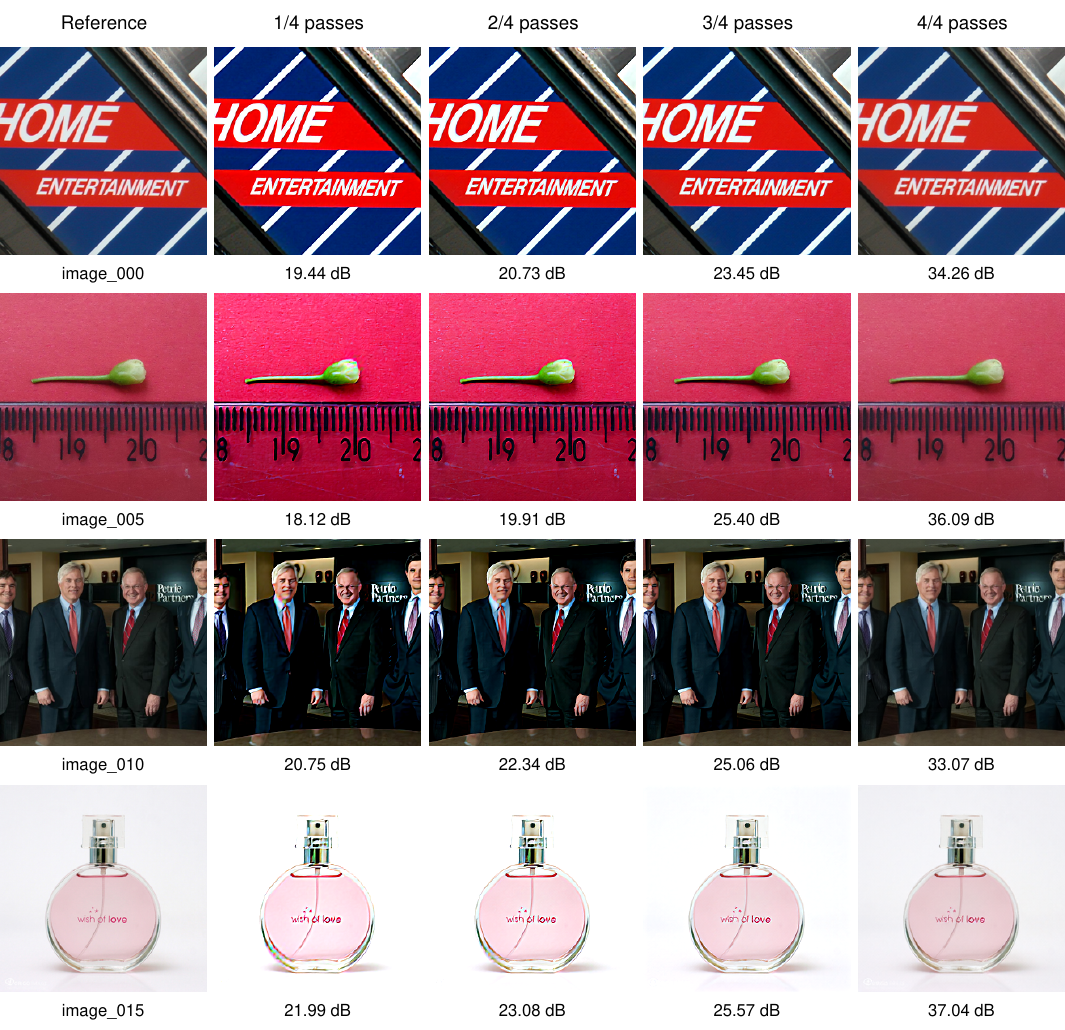}
  \caption{\textbf{Four cases selected at evenly spaced indices in the recorded order.} Rows use zero-based indices 0, 5, 10, and 15, selected without ranking reconstruction quality. Only the encoder f8 prefix changes; the downstream network retains its trained schedule. Columns show the reference and one through four retained passes, with per-image PSNR. All panels use the same RGB mapping and clipping, with no per-panel contrast adjustment. The arrays are measured reconstructions, not illustrative predictions.}
  \label{fig:loop_cases}
\end{figure}

\paragraph{Shared weights do not imply equivalent loop positions.}
All 224 separately applied single-pass bypasses reduce PSNR relative to their image's full schedule. At encoder f8, bypassing passes 1--4 produces mean losses of 3.66, 9.87, 7.54, and 7.78 dB. At decoder f2, the losses are 8.44, 11.42, 13.16, and 12.92 dB. The second encoder f8 pass and third decoder f2 pass are the most sensitive positions within their respective stages on all 16 images. The largest loss is therefore not assigned to the last pass, despite the large final-step gain in the prefix experiment. These observations support position-dependent computation under shared weights. They do not isolate loop conditioning from the evolving hidden state, and bypass losses cannot be added as independent contributions.

\begin{figure}[!t]
 \centering
 \includegraphics[width=\linewidth]{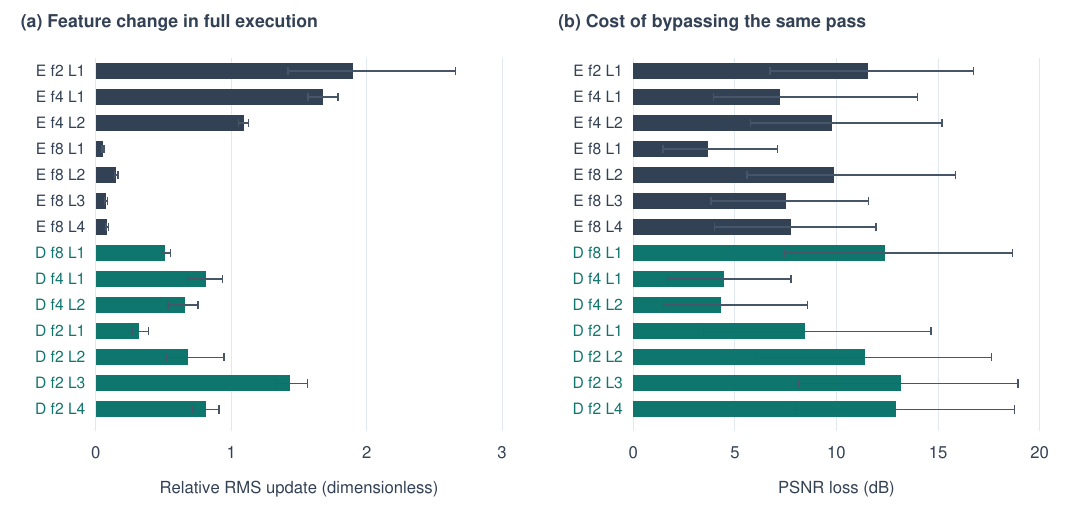}
 \caption{\textbf{Feature-change magnitude and reconstruction sensitivity measure different properties.} Each row identifies the same pass in a full forward trace (left) and an independent bypass experiment (right). E/D denote encoder/decoder, and L is the one-based loop index. Bars show means over 16 images; whiskers show the observed minimum and maximum, not confidence intervals. In particular, small encoder f8 updates can have large downstream consequences. These are paired measurements on the existing subset, not additional model evaluations.}
 \label{fig:loop_update}
\end{figure}

\paragraph{Small feature updates can be consequential.}
We pair the normal forward trace with the bypass measurement for each image and pass. Define the relative update as
\begin{equation}
 r_{s,t}=\frac{\operatorname{RMS}(h_{s,t+1}-h_{s,t})}
 {\max(\operatorname{RMS}(h_{s,t}),10^{-8})},
 \label{eq:relative_update}
\end{equation}
where $h_{s,t}$ is the pre-injection state, so the numerator includes injection and core processing. Figure~\ref{fig:loop_update} shows that a small change in this norm does not imply a dispensable pass. The fourth encoder f8 pass changes the feature by only 7.47--9.36\% across images, yet bypassing it loses 4.00--11.94 dB PSNR (mean 7.78 dB). Thus even updates below a relative threshold of 0.1 can be important to reconstruction in this checkpoint. The dimensionless norm is a description of the current feature coordinates, not a calibrated estimate of output error or a validated stopping criterion.

\paragraph{Output-range recovery, not established early-exit refinement.}
The qualitative cases show recognizable structure before the last pass, but also substantial color and contrast distortion (Figure~\ref{fig:loop_cases}). The final pass raises mean PSNR by 7.78 dB at encoder f8 and 12.92 dB at decoder f2. Crucially, decoder f2 prefixes of length one, two, and three produce out-of-range RGB values at 96.72\%, 93.44\%, and 68.34\% of pixels on average, versus 6.14\% for the full schedule. Before clipping, their mean PSNR values are $-1.04$, 3.42, and 14.45 dB, compared with 32.25 dB for the complete model (Table~\ref{tab:loop_clipping}). Clipping therefore masks substantial early-exit distortion rather than creating the full-depth advantage: the complete model's mean clipping benefit is only 0.028 dB. The evidence demonstrates useful recurrent computation and sensitivity to the trained schedule, but does not establish that intermediate states are calibrated for early exit, that iteration converges, or that additional untrained loops would help.

\FloatBarrier

\subsection{Multi-Resolution Reconstruction}

\begin{table}[t]
\centering
\caption{Joint Multi evaluation at the Stage-2 220k checkpoint (approximately 5.5 nominal Stage-2 epochs; Stage 1 excluded). All three rows use one checkpoint and one shared encoder traversal, followed by decoding from the indicated latent interface.}
\label{tab:multi_results}
\begin{papertabular}{lccccc}
\toprule
Latent & Shape at $256^2$ & rFID $\downarrow$ & PSNR $\uparrow$ & LPIPS $\downarrow$ & SSIM $\uparrow$ \\
\midrule
8x  & $16\times32\times32$ & 0.768 & 29.156 & 0.0868 & \textbf{0.8398} \\
16x & $64\times16\times16$ & \textbf{0.718} & \textbf{29.197} & \textbf{0.0856} & 0.8395 \\
32x & $128\times8\times8$  & 1.436 & 27.107 & 0.1413 & 0.7728 \\
\bottomrule
\end{papertabular}
\end{table}

The joint Stage-2 220k checkpoint reconstructs from all three latent interfaces (Table~\ref{tab:multi_results}).  Its 8x and 16x interfaces have similar paired reconstruction quality, while 32x reduces PSNR, LPIPS quality, and SSIM.  These operating points demonstrate one backbone with several latent interfaces, not superiority over three separately trained tokenizers.

\subsection{Downstream Diffusion}

We train class-conditional LightningDiT models on normalized \texttt{f8d16} latents using the FasterDiT and VA-VAE pipeline~\cite{yao2024fasterdit,yao2025vavae}. Both B/2 and XL/2 use the Stage-2 610k CNN tokenizer (approximately 15 nominal Stage-2 epochs, after Stage-1 pretraining) and are evaluated on 50,000 generated images without classifier-free guidance (CFG). Class conditioning remains active. The tokenizer is frozen throughout generator training.

\begin{table}[t]
\centering
\caption{Class-conditional generation on CNN LoopVAE latents, without CFG. Both generators use the Stage-2 610k tokenizer, are evaluated after 200k generator updates, and use 50k generated samples for FID. The XL/2 diffusion-model run corresponds to 80 training epochs; this is separate from the tokenizer's two-stage budget.}
\label{tab:dit_results}
\begin{papertabular}{lcccc}
\toprule
Generator & Updates & CFG & Samples & FID-50k $\downarrow$ \\
\midrule
LightningDiT-B/2 & 200k & off & 50k & 39.906 \\
LightningDiT-XL/2 & 200k & off & 50k & 18.198 \\
\bottomrule
\end{papertabular}
\end{table}

The B/2 and XL/2 runs reach FID-50k values of 39.906 and 18.198, respectively, demonstrating that the learned latents support downstream diffusion. These runs do not isolate tokenizer quality from generator capacity or training cost: a matched alternative tokenizer is not included, and equal update counts alone do not imply equal training budgets. We therefore report the generation results as an application of the tokenizer rather than a controlled scaling comparison.

\subsection{Parameters, Arithmetic, and Runtime}
\label{sec:resources}

Table~\ref{tab:efficiency} reports a benchmark on a single GPU with 96\,GB of device memory at $256^2$ resolution with bfloat16 execution, batch size eight, ten warm-up iterations, and 100 timed iterations. The evaluator instantiates encoder/decoder pairs from configurations and runs encoding, posterior sampling, and decoding on synthetic inputs. This is an architectural benchmark, not an accuracy-matched comparison of pretrained checkpoints. The label \emph{Flux-style} denotes a local CNN reference configuration rather than an official Flux performance measurement. Parameter counts are specific to the profiled configurations; in particular, the 32.46M V3 ViT is distinct from the reported 34M reconstruction model.

\begin{table}[t]
\centering
\caption{Architectural benchmark on a single GPU with 96\,GB of device memory. Parameters cover the encoder and decoder; MACs and FLOPs are estimates per image, with one MAC counted as two FLOPs. Time/image is batch latency divided by eight, not batch-one latency. Memory is peak allocated MiB (the log labels it MB).}
\label{tab:efficiency}
\begin{papertabular}{llrrrrrr}
\toprule
Scale & Model & Params (M) & MACs (G) & FLOPs (G) & Images/s & ms/image & MiB \\
\midrule
8x & Looped CNN V3 & 29.07 & 700.92 & 1401.84 & 28.5 & 35.13 & 1729.3 \\
8x & Looped ViT V3 & 32.46 & 959.89 & 1919.79 & 28.0 & 35.72 & 1359.6 \\
8x & CNN (Flux-style) & 83.82 & 447.62 & 895.25 & 81.0 & 12.34 & 1302.8 \\
8x & ViT-B & 172.01 & 213.01 & 426.02 & 270.4 & 3.70 & 521.2 \\
8x & ViT-L & 607.08 & 722.10 & 1444.21 & 73.8 & 13.54 & 1379.4 \\
\midrule
16x & Looped CNN V3 & 32.67 & 470.56 & 941.12 & 42.3 & 23.67 & 1746.3 \\
16x & Looped ViT V3 & 36.06 & 587.94 & 1175.88 & 39.4 & 25.38 & 1369.1 \\
16x & CNN & 69.83 & 195.06 & 390.13 & 133.4 & 7.50 & 956.5 \\
16x & ViT-B & 171.51 & 46.18 & 92.36 & 731.3 & 1.37 & 428.4 \\
16x & ViT-L & 606.41 & 161.43 & 322.86 & 268.7 & 3.72 & 1249.2 \\
\midrule
32x & Looped ViT V5 & 208.31 & 2629.46 & 5258.91 & 13.9 & 71.81 & 2966.9 \\
32x & CNN & 99.49 & 191.86 & 383.71 & 132.9 & 7.53 & 1028.9 \\
\bottomrule
\end{papertabular}
\end{table}

At 8x, the looped CNN stores 65.3\% fewer parameters than the Flux-style reference, but uses $1.57\times$ its estimated MACs and takes $2.85\times$ its batch latency (281.05 versus 98.73 ms).  Its peak allocated memory is also higher.  The fixed-resolution ViT-B is much faster despite storing more parameters; its execution graph differs substantially from recurrent high-resolution processing.  Fewer weights therefore do not imply fewer operations, lower activation memory, or faster inference.

\paragraph{Where repeated computation concentrates.}
For a fixed-width CNN core, the convolutional work is proportional to $KL_sH_sW_s$. At $256^2$ input resolution, encoder f2/f4/f8 and decoder f8/f4/f2 contribute proportionally to $16:8:4:1:8:64$, respectively. The four passes at the finest decoder stage therefore account for $64/101=63.4\%$ of this spatially linear core work. This is an analytical decomposition of the specified graph, excluding transitions, heads, and conditioning overhead, not a measured full-model FLOP breakdown. It explains why the 28-block count alone obscures an important cost: the same block is much more expensive at higher spatial resolution. The sensitivity of decoder f2 to truncation also shows that removing this work from the existing checkpoint is not quality-neutral.

The 32x V5 benchmark includes all stored encoder/decoder parameters and multi-head encoder computation, using the 32x decoder entry. The looped CNN V5 failed to instantiate and has no measurement. Different latent shapes and loop schedules make the 16x/32x rows an uncontrolled compression comparison. Arithmetic estimates cover convolution, linear, and attention matrix products, but omit some normalization, activation, and elementwise work. Runtime is measured independently on the same device for all models; timings need not transfer to other GPUs with the same memory capacity.

\FloatBarrier
\section{Discussion and Limitations}
\label{sec:analysis}

\paragraph{Parameter sharing is a capacity choice, not an inference shortcut.}
The central result is that a small core can support a deep visual hierarchy without assigning different weights to every scale and pass. The Transformer ablation provides the most direct evidence: global sharing remains competitive in paired fidelity at one seventh the stored core depth, while unshared processing improves LPIPS. The final CNN result demonstrates a useful parameter--quality operating point, but the unmatched baseline recipes do not establish a causal advantage of recurrence. Parameter storage may motivate this design; the measured arithmetic, latency, and activation memory do not support an execution-efficiency claim.

\paragraph{Useful recurrence need not resemble convergence.}
The intervention study shows that the trained schedule matters, including passes whose relative feature changes are small. Completing a stage improves reconstruction on the examined images, but severe raw-output errors under truncation make intermediate quality different from full-schedule quality. These observations are consistent with a depth-dependent transformation rather than interchangeable refinement steps. They neither establish fixed-point convergence nor isolate the roles of loop conditioning and evolving hidden states. In particular, a bypass tests dependence within the trained network; it does not predict the performance of a shallower model trained from scratch.

\paragraph{The high-resolution schedule is a concrete optimization target.}
The finest decoder stage dominates the spatially linear CNN core work, yet is also sensitive to truncation. This combination suggests that reducing deployment cost requires adapting training, not simply omitting passes after training. A useful next test is a matched-budget schedule comparison that reallocates passes between coarse and fine scales. Intermediate-output supervision or distillation~\cite{goyal2026elt} could separately test whether a shared tokenizer can support reliable early exit; the current measurements do not answer that question.

\paragraph{Evidence boundaries.}
The sharing comparison is a single-run, non-adversarial Transformer experiment; seed variation, identical data order, and separate effects of scale embeddings, loop embeddings, and input injection remain unresolved. The loop study is a targeted TokBench case analysis, not a full benchmark evaluation, and does not measure text-recognition accuracy, face similarity, or LPIPS. Reconstruction, diagnostics, generation, and profiling have distinct checkpoint or configuration scopes, documented in the appendix. Multi-resolution heads change channel capacity as well as spatial size and lack separately trained controls. Both generation results use FID-50k, but a matched alternative tokenizer and a complete evaluation-sampler record are not available. Higher-resolution transfer, video, and text-conditioned generation remain untested.

A controlled tokenizer comparison, broader loop-intervention evaluation, and repeated sharing or schedule runs would test whether the observed tradeoff persists across checkpoints and data distributions. These extensions complement the present evidence that globally shared processing can support useful visual tokenization while retaining a substantial dependence on its trained execution schedule.

\section{Conclusion}

LoopVAE separates resolution-changing transitions from a processing core shared within and across spatial scales. A four-block core realizes 28 block applications per branch, with CNN and Transformer implementations and single- or multi-resolution latent interfaces. The 29M-parameter CNN reaches 0.28 rFID and 32.54 dB PSNR on ImageNet-256 under an approximately 30-epoch two-stage training budget; a separate sharing ablation finds a metric-dependent tradeoff between global reuse and unshared capacity.

The diagnostics clarify what parameter reuse does and does not buy. Trained passes contribute to reconstruction even when their relative feature changes are small, but truncated outputs are not reliably calibrated. High-resolution recurrence also costs substantial arithmetic and runtime despite the small stored core. These results establish recurrent depth across scales as a viable parameter-sharing organization for visual tokenization and identify schedule design and intermediate-output training as concrete directions for improving its execution tradeoff.

\begingroup
\interlinepenalty=10000
\bibliographystyle{plainnat}
\setlength{\bibsep}{3pt}
\bibliography{main}
\endgroup

\clearpage
\appendix
\section{Implementation and Reproducibility Details}

\subsection{LoopVAE-Single Configuration}
\label{sec:training_budget}

Table~\ref{tab:appendix_single_config} records the configuration of the main convolutional checkpoint.  Encoder and decoder use different parameters.  The decoder traverses scales from 8x to 2x; its loop list is indexed in traversal order, so the stored list $[1,2,4]$ corresponds to one, two, and four repeated passes during that reverse traversal.

\begin{table}[h]
\centering
\caption{Main LoopVAE-Single configuration.}
\label{tab:appendix_single_config}
\begin{papertabular}{lclc}
\toprule
Architecture & Value & Training & Value \\
\midrule
Input resolution & $256\times256$ & Precision & bfloat16 mixed \\
Hidden width & 384 & Devices & 8 GPUs \\
Core blocks per branch & 4 & Per-device batch & 4 \\
Expansion ratio & 4 & Gradient clipping & 1.0 \\
Spatial kernel & $7\times7$ depthwise & Stage-2 learning rate & $5\times10^{-5}$ \\
Encoder traversal & 2x, 4x, 8x & KL weight & $10^{-6}$ \\
Decoder traversal & 8x, 4x, 2x & Latent noise std. & 0.05 \\
Loop list per branch & $[1,2,4]$ & LPIPS weight & 1.0 \\
Latent shape & $16\times32\times32$ & GAN activation step & 5,000 \\
Total parameters & 29.07M & GAN weight multiplier & 0.5 \\
\bottomrule
\end{papertabular}
\end{table}

Stage 1 trains without GAN for 600k steps by default. Stage 2 initializes from the Stage-1 autoencoder weights and continues with GAN training; all Stage-2 checkpoint step labels exclude Stage 1. The main CNN and ViT reconstruction results use a nominal total budget of approximately 30 epochs, split approximately equally between the two stages.

For a global batch of $8\times4=32$ and approximately 1.28M ImageNet training images, one epoch contains approximately 40k data batches. Under the reported data-iteration convention, 600k and 610k steps correspond to approximately 15.0 and 15.2 epochs, respectively; 220k corresponds to approximately 5.5 epochs. We round the main two-stage budget to approximately 30 epochs. These are nominal data-exposure estimates, not an exact reconstruction of checkpoint counters. In the available manual-optimization code, Lightning's raw \texttt{global\_step} can count both autoencoder and discriminator updates; exact exposure would require the corresponding epoch or data-batch logs. We retain the stage-local checkpoint labels to identify the evaluated models.

Stage 2 of the main CNN uses the frozen \texttt{facebook/dinov2-small} feature extractor and a trainable discriminator head.  The local recipe name \texttt{S\_16} is an alias, not its patch size: the backbone uses $14\times14$ patches.  The generator's adversarial term activates at the configured Stage-2 global-step threshold of 5,000, after discriminator warm-up; this threshold is separate from the Stage-1 600k-step duration.  We use hinge loss, differentiable augmentation probability 1.0 with cutout 0.2, and lazy R1 regularization with weight 10 every 16 discriminator updates.  The adaptive generator coefficient is the ratio of reconstruction and adversarial gradient norms at the output layer, clipped at $10^4$ and multiplied by 0.5.

\subsection{LoopVAE-Multi Configuration}

The five encoder scales are 2x, 4x, 8x, 16x, and 32x.  The shared core contains six Transformer blocks at width 768, with 12 attention heads, and uses loop schedule $[1,1,2,3,6]$.  Attention uses $8\times8$ windows at the two finest scales and global connectivity at the remaining scales.  Posterior heads are attached to scale indices 2, 3, and 4.  Per-interface losses are weighted by 0.5, 1.0, and 1.5; each uses KL weight $10^{-6}$.  Stage 2 uses a convolutional discriminator with depth setting three and adaptive adversarial weight multiplier 0.2, rather than the main CNN run's DINO-based discriminator.

\Needspace{4\baselineskip}
The multi-entry decoder has a separate input projection for each latent channel count and loop list $[6,3,2,1,1]$ in 32x-to-2x traversal order.  Starting from 8x skips the 16x and 32x synthesis stages; starting from 16x skips only 32x; starting from 32x executes the complete reverse hierarchy.  These entry points require 24, 42, and 78 core block applications, respectively, all using the same six stored blocks.

\subsection{Evaluation Protocol}

For reconstruction, all validation images are encoded with the posterior mode and decoded once.  PSNR, LPIPS, and SSIM are averaged over paired images after mapping tensors to the same image range.  rFID uses the same 50k ImageNet validation images as both real-image identities and reconstruction sources.

Reported scores are individual checkpoint evaluations, not averages across training seeds. Table~\ref{tab:single_results} gives the approximately 30-epoch, two-stage \texttt{f8d16} results, while the sharing ablation compares the Transformer-based global, scale-wise, and fully unshared variants at Stage 1, step 400k, without adversarial training. The ablation reports paired reconstruction metrics. Its run records do not establish identical data ordering or seed-level uncertainty.

The final reconstruction evaluations, 610k CNN diagnostics and generation runs, joint 220k Multi evaluation, and synthetic profiling have separate checkpoint or configuration scopes. The final reconstruction checkpoint identifiers have not been linked to the diagnostic checkpoints; results are therefore not combined across these evaluations. Latent interpolation is not evaluated in this paper.

\subsection{Loop Intervention Protocol and Output Range}
\label{sec:loop_protocol}

The diagnostic in Section~\ref{sec:loop_dynamics} uses the raw encoder/decoder weights of \texttt{abla\_looped\_f8\_v3\_cnn\_stage2\_step610k.ckpt}, loaded strictly without an EMA substitution. The measurement manifest records the checkpoint SHA256 beginning \texttt{81b02d86d5b9fab9}, the resolved configuration, source-code hashes, and each input image's identity and hash. The test images originate from TokBench~\cite{wu2025tokbench}, whose emphasis on faces and text motivates the case study. Sixteen unique source-file hashes were verified. Within the local test-image collection, the first input is \texttt{img\_105.jpg}, already examined in a pilot; the other inputs are the first 15 remaining paths in sorted order after path deduplication. This is a targeted, non-random selection, not an outcome-ranked sample. Figure~\ref{fig:loop_cases} uses evenly spaced zero-based indices 0, 5, 10, and 15 from the recorded order.

Images undergo EXIF orientation correction, RGB conversion, bicubic resizing of the shorter side to 256, and a centered $256\times256$ crop. The reference is retained in float32 before casting model inputs to bfloat16. Posterior mode is used throughout. Each image yields 14 prefix records and 14 separately applied single-pass bypass records, plus a normal forward trace of 14 passes. Hooks preserve the baseline output exactly; full-prefix arrays also match the baseline. Bypasses execute the pass before replacing its output, so their runtime cannot be used to claim computational savings.

Let $u=(\hat{x}_{\rm raw}+1)/2$ be the mapped RGB output and $y\in[0,1]$ the reference. Standard diagnostic PSNR and SSIM use $\operatorname{clip}(u,0,1)$. Raw PSNR instead uses $-10\log_{10}(\operatorname{MSE}(u,y))$ with data range one; either PSNR is capped at 120 dB for an exact match. SSIM uses a valid $11\times11$ Gaussian window with $\sigma=1.5$. LPIPS and rFID are not measured on this subset. The out-of-range pixel fraction counts a pixel if any channel is strictly below zero or above one. Clipping displacement is $\operatorname{mean}|u-\operatorname{clip}(u,0,1)|$. Raw extrema, channel- and pixel-level fractions, and raw errors are saved as scalar diagnostics, whereas saved reconstruction arrays contain clipped RGB.

\begin{table}[ht]
\centering
\caption{Decoder f2 prefix diagnostic on 16 images. Values are means over images; the full model is the four-pass row. Clipping displacement is an RGB mean absolute change, not a reconstruction error against the reference. A negative raw PSNR is possible because raw outputs are not restricted to the unit interval.}
\label{tab:loop_clipping}
\begin{papertabular}{crrrr}
\toprule
Retained passes & Clipped PSNR & Raw PSNR & Out-of-range pixels & Clipping displacement \\
\midrule
1 & 14.54 & $-1.04$ & 96.72\% & 0.86934 \\
2 & 15.26 & 3.42 & 93.44\% & 0.47082 \\
3 & 19.36 & 14.45 & 68.34\% & 0.08287 \\
4 (full) & 32.28 & 32.25 & 6.14\% & 0.00019 \\
\bottomrule
\end{papertabular}
\end{table}

Out-of-range fraction alone does not measure the severity of distortion. The white-background vehicle image has 55.73\% out-of-range pixels at full depth but only 0.00169 mean clipping displacement and a 0.046 dB clipping benefit. Conversely, some one-pass decoder outputs extend as far as $-2.67$ and $4.09$ in mapped RGB. Magnitude and fraction must therefore be interpreted together. The full model averages 32.279 dB clipped and 32.251 dB raw PSNR across this subset.

Statistics use images as the unit, not loops or pixels. Prefix monotonicity is checked per image and stage with tolerance $10^{-6}$; one-pass stages have no adjacent comparison. All 64 multi-pass image--stage pairs are monotone in clipped PSNR, raw PSNR, and SSIM. Excluding the previously inspected pilot leaves 60/60 monotone pairs and 210/210 bypasses with lower PSNR. This sensitivity check reduces dependence on the pilot but does not make the remaining convenience sample random or establish population-level monotonicity.

The trace analysis joins records by image identity, branch, scale, and loop index, yielding 224 unique trace--bypass pairs. RMS is taken over the entire feature tensor, with the pre-injection state as reference. Feature-update ranges describe variability across images rather than uncertainty intervals. All qualitative panels use the measured, clipped RGB arrays with a common display mapping; no per-image contrast normalization is applied. These measurements characterize sensitivity within the trained checkpoint and do not establish semantic specialization or population-level monotonicity.

\subsection{Generation and Profiling Details}
\label{sec:additional_eval}

\paragraph{Generation records.}

Both generators use the Stage-2 610k CNN tokenizer and are evaluated after 200k generator updates using 50,000 generated images, with class conditioning and without CFG. The XL/2 diffusion-model run corresponds to 80 epochs, independently of the tokenizer's two-stage training budget. Their unrounded FID values are 39.90569705879807 for B/2 and 18.19847535025633 for XL/2; Table~\ref{tab:dit_results} reports three decimals. A saved training configuration differs from the reported evaluation in tokenizer path and guidance setting, so it is not treated as a complete record of the evaluation sampler. No comparison to a matched alternative tokenizer is available.

\paragraph{Profiling protocol.}
The evaluator casts parameters and buffers to bfloat16, times 100 batched encoder--sample--decoder passes after ten warm-ups, and synchronizes CUDA before and after timing. Arithmetic is counted with a batch-one input and reported per image. Memory uses \texttt{max\_memory\_allocated}/$1024^2$, hence MiB. The measurement excludes data loading, diffusion sampling, and discriminator execution. The 32x V5 row selects the 32x posterior and decoder entry but retains the multi-head encoder computation. All reported execution measurements follow the profiling protocol in Section~\ref{sec:resources}.

\end{document}